\documentclass{article}

\usepackage{subfiles}
\usepackage{arxiv}

\usepackage[utf8]{inputenc} 
\usepackage[T1]{fontenc}    
\usepackage{hyperref}       
\usepackage{url}            
\usepackage{booktabs}       
\usepackage{amsfonts}       
\usepackage{nicefrac}       
\usepackage{microtype}      
\usepackage{lipsum}
\usepackage{graphicx}
\usepackage{amsmath}  
\usepackage{mathrsfs}
\usepackage{multirow}
\usepackage{physics}
\usepackage{amssymb,amsfonts,amsthm}
\usepackage{bm}
\graphicspath{ {./images/} }

\usepackage{lipsum}                     
\usepackage{xargs}                      
\usepackage[colorinlistoftodos,prependcaption,textsize=tiny]{todonotes}
\newcommandx{\unsure}[2][1=]{\todo[linecolor=red,backgroundcolor=red!25,bordercolor=red,#1]{#2}}
\newcommandx{\change}[2][1=]{\todo[linecolor=blue,backgroundcolor=blue!25,bordercolor=blue,#1]{#2}}
\newcommandx{\info}[2][1=]{\todo[linecolor=OliveGreen,backgroundcolor=OliveGreen!25,bordercolor=OliveGreen,#1]{#2}}
\newcommandx{\improvement}[2][1=]{\todo[linecolor=Plum,backgroundcolor=Plum!25,bordercolor=Plum,#1]{#2}}
\newcommandx{\thiswillnotshow}[2][1=]{\todo[disable,#1]{#2}}
\newcommand\blfootnote[1]{%
  \begingroup
  \renewcommand\thefootnote{}\footnote{#1}%
  \addtocounter{footnote}{-1}%
  \endgroup
}
\title{A Systematic Approach to Featurization for Cancer Drug Sensitivity Predictions with Deep Learning}
\author{
  Austin Clyde$^{1,2}$
  \And
  Tom Brettin$^{2}$
  \And 
  Alexander Partin$^{2}$
  \And 
  Maulik Shaulik$^{2}$
  \And
  Hyunseung Yoo$^{2}$
  \And
  Yvonne Evrard$^{3}$
  \And
  Yitan Zhu$^{2}$
  \And 
  Fangfang Xia$^{2}$
  \And
  Rick Stevens$^{1,2,\ddagger}$
}

\begin{document}

\maketitle

\blfootnote{$^1$ Department of Computer Science, University of Chicago, Chicago, IL, United States.}
\blfootnote{$^2$ Computing, Environment, and Life Sciences, Argonne National Laboratory, Lemont, IL, United States.}
\blfootnote{$^3$ Applied Development and Research Directorate, Frederick National Laboratory for Cancer Research, Frederick, MD, United States.}
\blfootnote{$\ddagger$ Primary contact PI Rick Stevens stevens@anl.gov}
\begin{abstract}
    By combining various cancer cell line (CCL) drug screening panels, the size of the data has grown significantly to begin understanding how advances in deep learning can advance drug response predictions. In this paper we train >35,000 neural network models, sweeping over common featurization techniques. We found the RNA-seq to be highly redundant and informative even with subsets larger than 128 features. We found the inclusion of single nucleotide polymorphisms (SNPs) coded as count matrices improved model performance significantly, and no substantial difference in model performance with respect to molecular featurization between the common open source MOrdred descriptors and Dragon7 descriptors. Alongside this analysis, we outline data integration between CCL screening datasets and present evidence that new metrics and imbalanced data techniques, as well as advances in data standardization, need to be developed.
\end{abstract}

\section{Introduction}
 The crux of precision oncology and virtual drug screening lies in the relationship between molecular structure and cancer genetics \cite{lo2007genetic,zhong1993relationship}. It is well studied that cancer genetics may play a role in drug efficacy even across similar cancer types \cite{candelaria2005genetic}. In order to understand this relationship, drug screening assays such as the NCI-60 Human Tumor Cell Line Screen were created to drive the precision oncology search for the link. Recently, deep learning has been applied to the problem as a means of modeling the interaction between cell genetics and drug properties \cite{manica2019towards, cortsciriano2019,xia2018,chang2018cancer}. We aim to explore the impact of featurization on a small subset of the possible feature space utilizing multiple combined dose response data. 
    
    Classical machine learning models such as Random Forest have been used in typing cancer genetics and even in single agent drug response models \cite{menden2013, rahman2019functional, papillon2013comparison}; however, most of these studies create single models based either predicting the response of a single cell given various drugs or predicting the activity of a drug over different cell lines. Few classical models have shown success across a wide range of CCLs and a diverse set of drugs in a single multitask model. Deep learning provides a more natural interface as higher dimensional datasets can be applied and reduced to a lower dimensional representation, where the choice of representation of the data is still important \cite{coates2011analysis}. 
    
    Prior featurization techniques were studied by Jang \textit{et. al}, where they systematically
    analyzed CCL drug sensitivity modeling as a classical machine learning task, by exploring the multitude of feature modalities, algorithms, prediction targets, and more \cite{jang2014systematic}. Given that they had a small set of drugs available between the studies (138 compound), the models Jang \textit{et. al} studied did not featurize the drugs in a continuous embedding; rather they were one-hot encoded whether present or not. Overall, in explaining the predictive variance between the over 110,000 models tested, they ranked the factors for variance among models, finding genetic features and the particular compound being predicted to be very explanatory, while the algorithm being considerably less important. Their results were further in line with other work on more general learning problems involving genetic features \cite{shi2010microarray}. 
    
    While many drug feature representations such as SMILES based encoders \cite{manica2019towards} or pharmacophore models \cite{skalic2019shape} are commonly used in deep learning models, we chose to compare only two variants of molecular descriptors: Mordred and Dragon7, which are some of the more common classical feature representations \cite{moriwaki2018mordred, mauri2006dragon}. Deep learning models may be moving toward more natural representations such as graphical representations; however, even the integrated dataset does not offer the sort of drug diversity seen in the papers effectively using these newer representations \cite{dutil2018towards}. 
    
    Across the recent works, the decision to use RNA-seq in conjunction with images, SMILES, or fingerprints was a choice dictated by the desired model architecture, explainability of the model, or resources available; however, these choices do not span the entire space of representations available for cells nor drugs. Cortés‑Ciriano and Bender found introducing convolutional neural networks (CNNs) for the drug image alongside Morgan fingerprints improved performance of cell line sensitivity predictions with a small effect size \cite{cortsciriano2019}. \textit{Manica et al.} compared fingerprints to the use of SMILES strings, showing an RMSE improvement from $0.122\pm 0.010$ their deep baseline to $0.104\pm0.005$ in IC50 predictions with the benefit of an explainability mechanism for improved mechanism of action (MOA) and perturbation study \cite{manica2019towards}. \textit{Xia et al.} found the use of proteome, expression, and molecular fingerprints to perform the best when studying combinations of drug pairs; however, proteome data is not available at scale of this study \cite{xia2018}. Chang et al. utilized a virtual docking technique in conjunction with CNNs to achieve $r^2>0.84$ on 244 drugs from GDSC \cite{chang2018cancer}. \textit{Rampášek et al.} used gene perturbation data in an autoencoder \cite{rampavsek2019dr}. The goal of these models is to provide insight into which drugs and cells should be tested in downstream drug development analysis with, for example, PDX models or organoids.  
    
    In this work, we outline multiple methods for preparation of pan-cancer data, training strategies, and analysis strategies of these models. As opposed to a hyperparameter optimization experiment which would provide a single predictive model, we aim to understand a small section of the overall feature space: primarily which genetic features are most effective and whether or not MOrdred open-source descriptors compare to Dragon7 descriptors. While performing hyperparameter optimization on each individual model is ideal, this is not tractable given the sheer number of simple discrete permutations available. In an attempt to approximate the optimization task, we also sweep over various model architectures, training strategies, and other learning hyperparameter optimization runs on models. Our results aim to provide guidelines for creating deep learning models for CCL models across studies, with the hope that more researchers continue to explore the space. 
    
    In order to facilitate downstream analysis, we are releasing all the code to generate the data from public sources and data collected from the model runs. Details on data preparation and deep learning methods utilized are in the materials and method section\footnote{Please see \url{https://github.com/DOE-NCI-Pilot1/CCLFeatureComparison}}. 
 
 \section{Results}
\begin{figure}[h]
        \centering
        \includegraphics[width=0.48\textwidth]{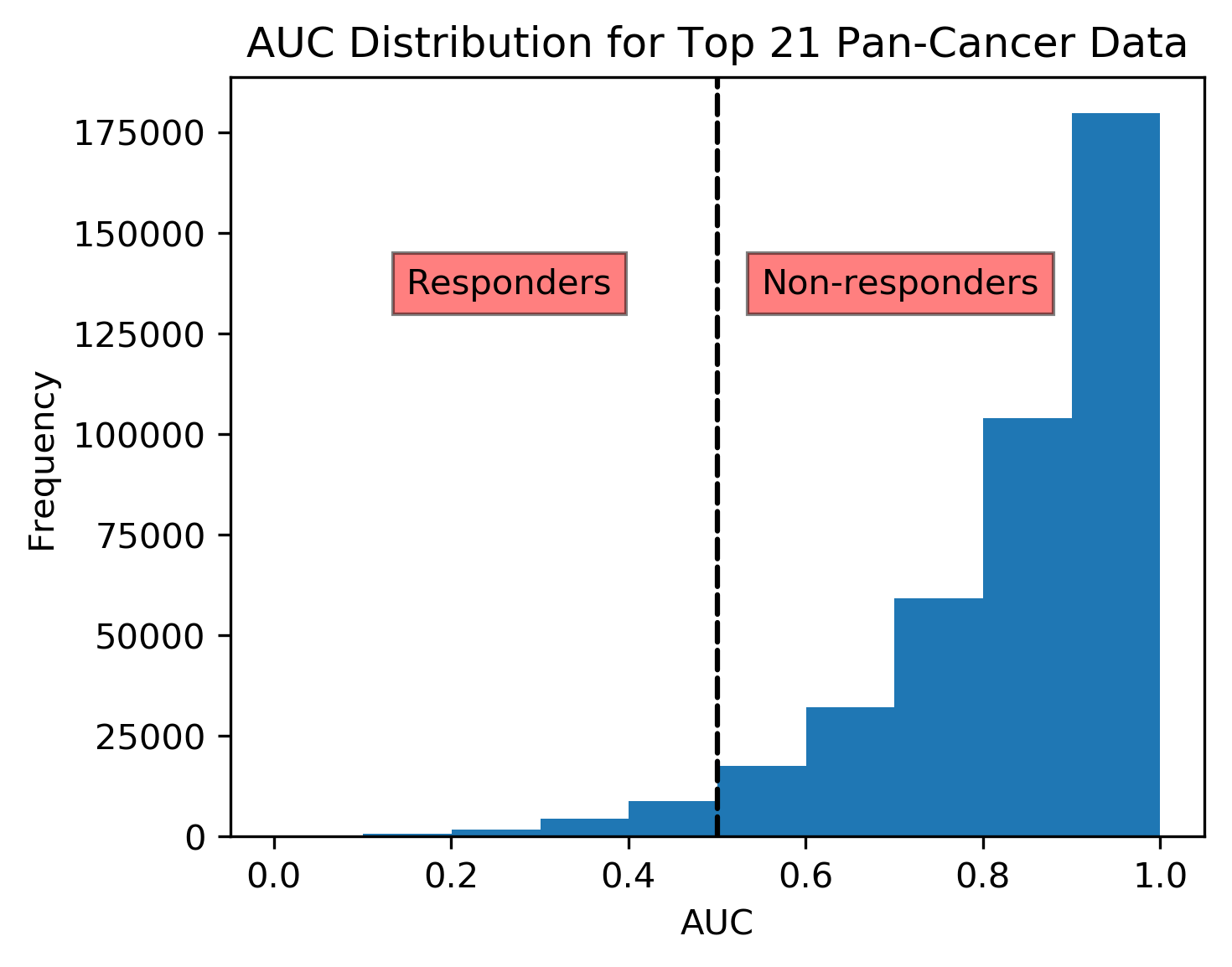}
        \includegraphics[width=0.48\textwidth]{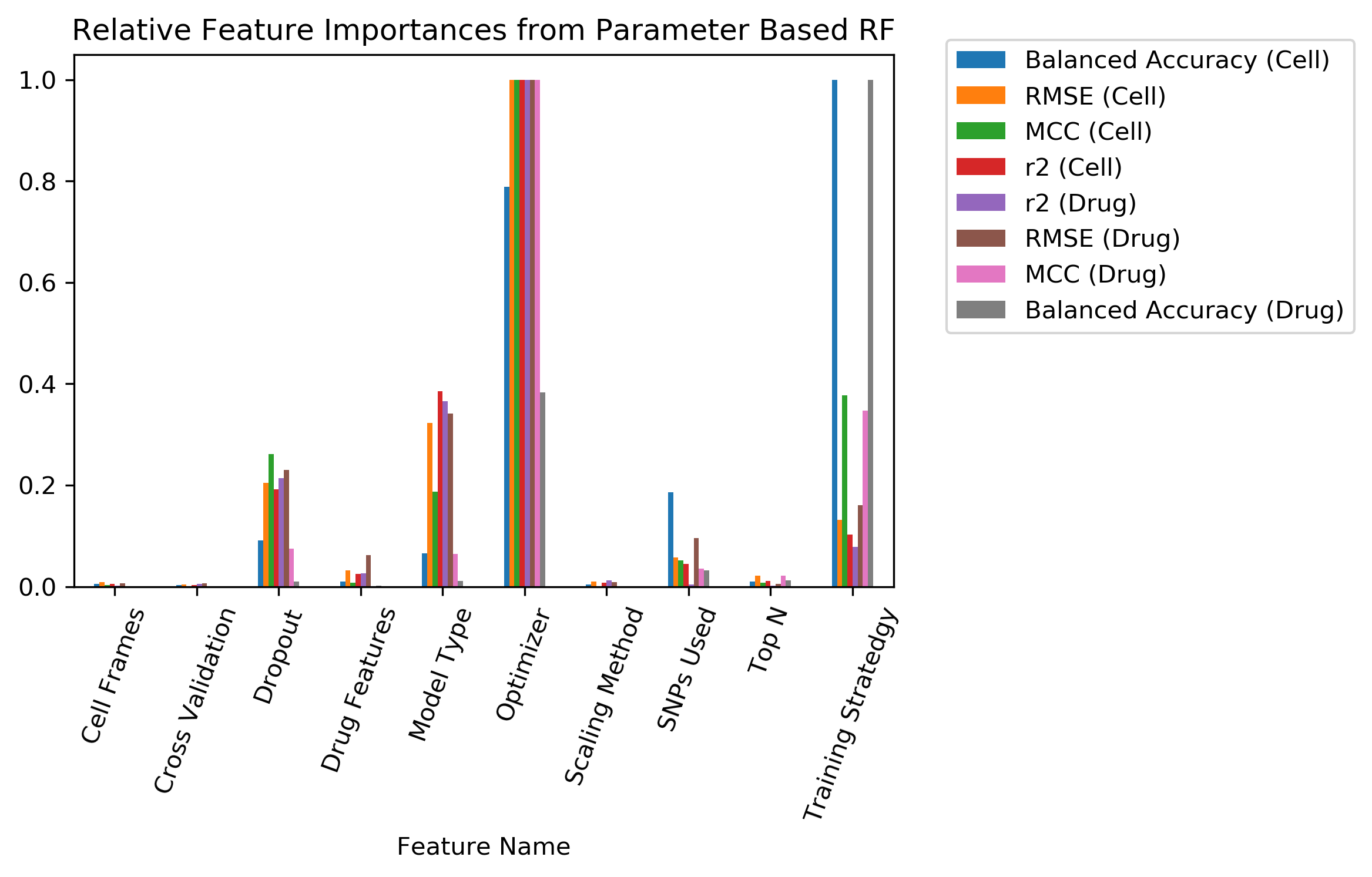}
        \caption{\textit{(Left)} The AUC distribution of pan-cancer data frame. With this scheme, the training distribution has 3.86\% responders. While our cutoff of 0.5 is arbitrary, learning to distinguish this slice along with a regression-based training strategy will prevent standard regression metrics from appearing much better than they are on the skewed portion of the dataset. \textit{(Right)}Relative feature importance for hyperparameters used in model training for predicting validation metric using decision trees. The $r^2$ for those models were $>0.9$ on cell validation metrics, and $>0.6$ for drug validation metrics. Optimizer, model type, and dropout were among the top three features, though the training strategy very important when predicting balanced accuracy.}
        \label{fig:feat_Important}
    \end{figure}

 We report on the results of models based on predictions on their assigned validation set and metrics. In terms of understanding the parameter space, we ran a series of classical machine learning predictors trained on the hyperparameter configuration to predict our four key metrics: RMSE, $r^2$, balanced accuracy, and MCC (Figure \ref{fig:feat_Important}). The results indicate standard deep learning hyperparameters to play the most important role in variance, rather than specific feature information, as optimizer, model architecture, training strategy, and dropout all are more predictive using SNPs or the featurization of drugs or cells.
    
    \subsection{Features Representation}
    The inclusion of SNPs improves performance on cell validation split models in RMSE, $r^2$, MCC, and balanced accuracy (Table \ref{tab:snp99}). There was no calculated metric in our test suite which reported a worse score when SNPs were included (at 100\% and 99\% percentiles). For the on cell validation method, the independent t-test shows significant improvement RMSE and $r^2$ scores with the inclusion of SNP features ($p<10^{-5}$), and less significant improvement for balanced accuracy and MCC ($p<0.17$ and $p<0.02$ respectively). For drug validation models, the RMSE and $r^2$ are again significantly different with the inclusion of SNP features ($p<0.003$ and $p<0.0005$) and not as significant for balanced accuracy and MCC ($p<0.01$ and $p<0.3$). 

    Scaling methods did not seem to have a large effect on the variance between models besides source scaling seemed to improve measures slightly (Table \ref{tab:scale99}. According to an ANOVA the null hypothesis of similarity among scaling methods is not rejected for either validation method ($p<0.14$). Cell frame splits did not have any noticeable differences amongst them (Table \ref{tab:cell99}). According to an ANOVA test for fit, there is no significant difference among cell validation methods $(p<0.31)$ and the drug validation method shows a possible difference but nothing significant $(p<0.06)$.
    
    The different chemical informatics package used to features the molecules did not have a significant effect on the validation scores (Table \ref{tab:drug99}). For on cell validation models, no reported metric rejects the hypothesis that all featurization are the same for $p < 0.09$; however, for on drug validation RMSE and $r^2$ scores are effected by featurization to an extent ($p<0.0001$). We applied post hoc analysis to these two metrics using John Turkey's HSD analysis which shows Dragon7 and MOrdred slightly different on $r^2$ and RMSE ($p<0.05$). The use of both MOrdred and Dragon7 however is significantly different from just using Dragon7 on both metrics ($p<0.01$).

\begin{table}[p]
\centering
\caption{Metric comparison at 99-percentile of grouped by the model validation strategy and the inclusion of SNPs. Rows do not represent a single model with those three metrics, rather a model exists with one of those metrics. A t-test for difference between group means shows a significant difference in $r^2$ scores $p=1.9e-18$ and $0.0005$ for cell and drug validation methods respectively.}\label{tab:snp99}
\begin{tabular}{@{}llllll@{}}
\toprule
Validation Method        & SNPs Included & RMSE  & $r^2$ score & Balanced Accuracy & MCC   \\ \midrule
\multirow{2}{*}{On Cell} & False         & 0.086 & 0.675       & 0.889             & 0.553 \\
                         & \textbf{True}          & 0.083 & 0.712       & 0.896             & 0.577 \\
\multirow{2}{*}{On Drug} & False         & 0.108 & 0.448       & 0.792             & 0.460 \\
                         & \textbf{True}          & 0.107 & 0.495       & 0.807             & 0.487 \\ \bottomrule
\end{tabular}
\end{table}
\begin{table}[p] 
\centering
\caption{Metric comparison at 99-percentile of grouped by the model validation strategy and the scaling method used. Rows do not represent a single model with those three metrics, rather a model exists with one of those metrics. }
\label{tab:scale99}
\begin{tabular}{@{}llllll@{}}
\toprule
Validation Method        & RNA-seq Scaling Method & RMSE  & $r^2$ score & Balanced Accuracy & MCC   \\ \midrule
\multirow{3}{*}{On Cell} & Combat                 & 0.084 & 0.708       & 0.893             & 0.568 \\
                         & None                   & 0.084 & 0.703       & 0.893             & 0.565 \\
                         & Source Scaled          & 0.084 & 0.710       & 0.891             & 0.568 \\
\multirow{3}{*}{On Drug} & Combat                 & 0.107 & 0.484       & 0.797             & 0.478 \\
                         & None                   & 0.110 & 0.470       & 0.795             & 0.474 \\
                         & Source Scaled          & 0.108 & 0.494       & 0.796             & 0.473 \\ \bottomrule
\end{tabular}
\end{table}
    \begin{table}[p]
\centering
\caption{Metric comparison at 99-percentile of grouped by the model validation strategy and the RNA-seq feature set used. Rows do not represent a single model with those three metrics, rather a model exists with one of those metrics.}
\label{tab:cell99}
\begin{tabular}{@{}llllll@{}}
\toprule
Validation Method        & RNA-seq Feature Set    & RMSE  & $r^2$ score & Balanced Accuracy & MCC   \\ \midrule
\multirow{3}{*}{On Cell} & lincs1000              & 0.084 & 0.708       & 0.895             & 0.567 \\
                         & oncogenes              & 0.084 & 0.706       & 0.891             & 0.574 \\
                         & oncogenes \& lincs1000 & 0.085 & 0.708       & 0.894             & 0.568 \\
\multirow{3}{*}{On Drug} & lincs1000              & 0.108 & 0.478       & 0.796             & 0.482 \\
                         & oncogenes              & 0.107 & 0.478       & 0.788             & 0.463 \\
                         & oncogenes \& lincs1000 & 0.108 & 0.487       & 0.810             & 0.478 \\ \bottomrule
\end{tabular}
\end{table}
\begin{table}[p]
\centering
\caption{Metric comparison at 99-percentile of grouped by the model validation strategy and the drug featurization method used. Rows do not represent a single model with those three metrics, rather a model exists with one of those metrics.}
\label{tab:drug99}
\begin{tabular}{@{}lllllll@{}}
\toprule
Validation Method        & Drug Featurization              & RMSE  & $r^2$ score & Balanced Accuracy & MCC   \\ \midrule
\multirow{3}{*}{On Cell} & Dragon7 Descriptors             & 0.084 & 0.706       & 0.894             & 0.570 \\
                         & Mordred Descriptors             & 0.084 & 0.708       & 0.889             & 0.568 \\
                         & Mordred and Dragon7 Descriptors & 0.084 & 0.709       & 0.896             & 0.567 \\
\multirow{3}{*}{On Drug} & Dragon7 Descriptors             & 0.108 & 0.478       & 0.793             & 0.466 \\
                         & Mordred Descriptors             & 0.107 & 0.473       & 0.811             & 0.481 \\
                         & \textbf{Mordred and Dragon7 Descriptors} & 0.107 & 0.491       & 0.793             & 0.471 \\ \bottomrule 
\end{tabular}
\end{table}
 
 \section{Discussion}
 We first comment on the overall model performance of a few models created in this study. Second, we evaluate prior claims that feature representation should matter as the primary explanation of variance between models. We finally evaluate the three significant findings to consider in future research when building DNN models on CCL sensitivity predictions such as the drug featurization, training strategies, and the inclusion of SNPs. 
    
\begin{figure}[p]
        \centering
        \includegraphics[width=0.45\textwidth]{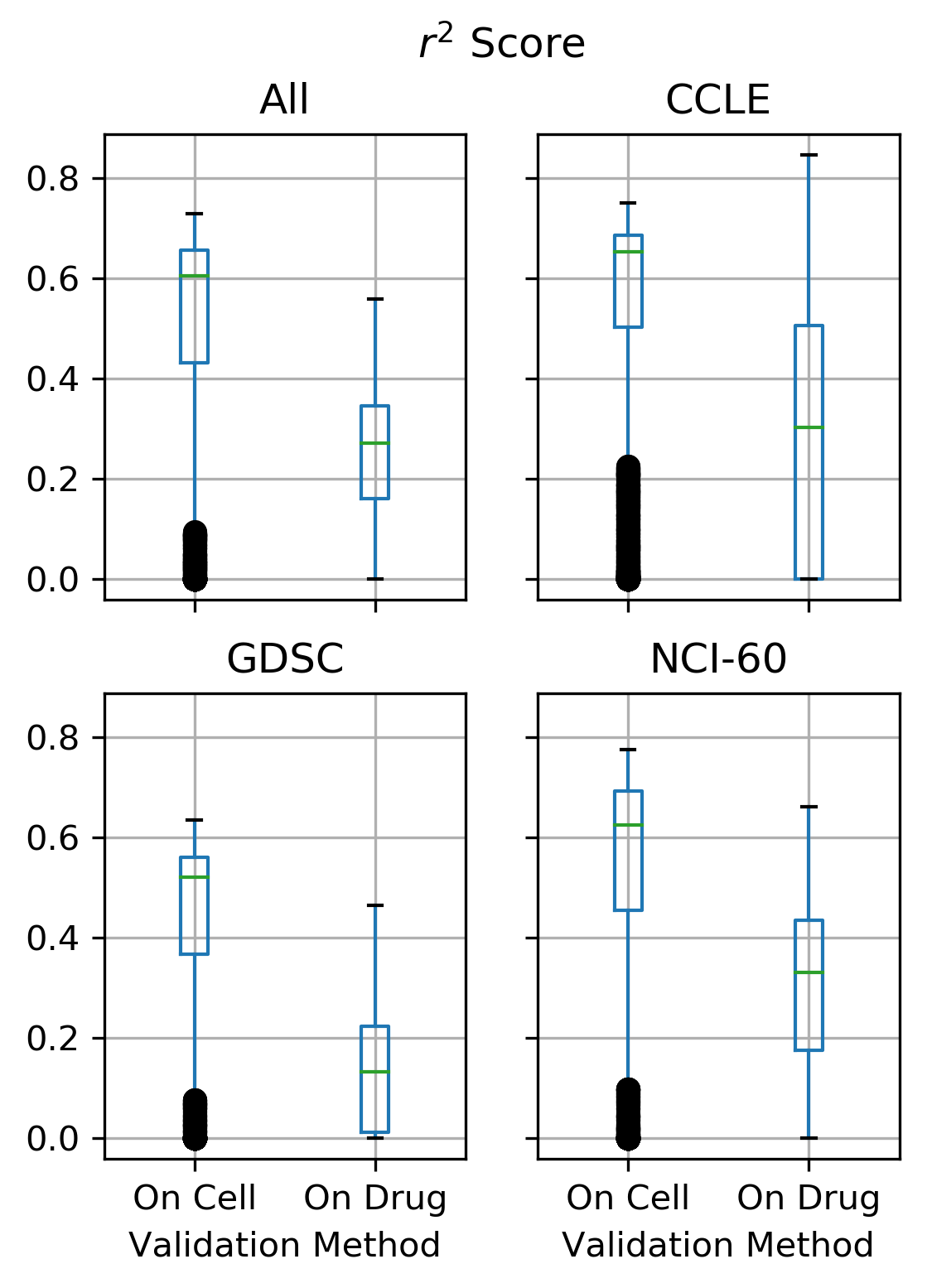}
        \includegraphics[width=0.45\textwidth]{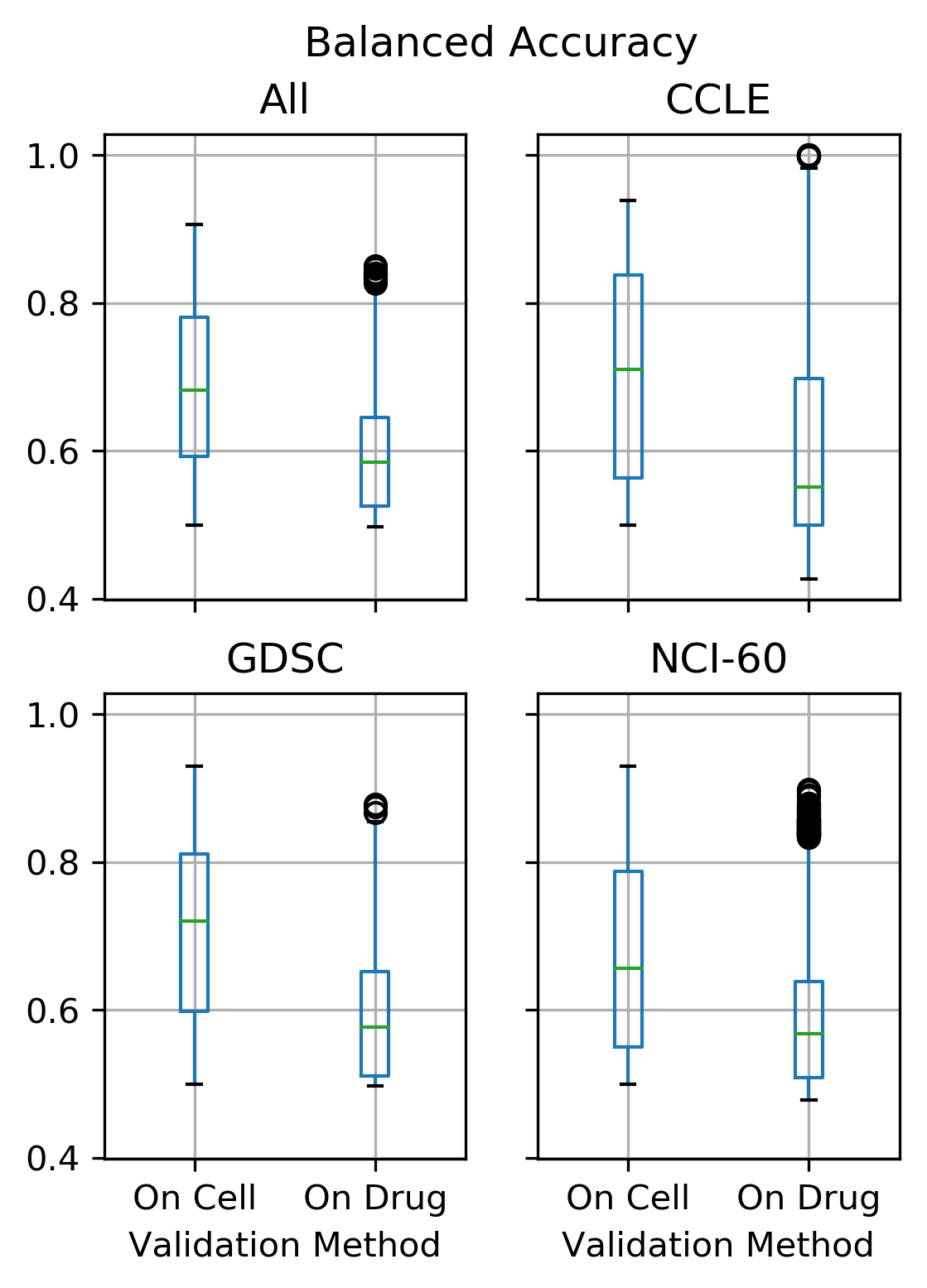}
        \caption{Comparison of the models' performances between validation methods over all converged trained models broken down by sample's originating study.}
        \label{fig:val}
    \end{figure}

    In general, we found the best models to perform quite well on the validation data in both regression and classification contexts. There are models with validation scores on unseen cells of $r^2>0.7$ and balanced accuracy $>90$\%, and there are models with validation on unseen drugs with $r^2>0.55$ and balanced accuracy $>84$\%. The cross-study performance of the two best cell validation models (Table \ref{tab:cross_study}) are on par with several other studies, though not more predictive on a single drug basis. The models in table \ref{tab:cross_study} are also top ranked if a simple average is taken across the cross-study measures as well. Models validated with unseen cells outperform models trained with unseen drugs in, however the performance varies across the studies where (Figure \ref{fig:val}). Given classical machine learning methods were generally capable of single drug predictions, we expected the cell validation performance to be better than the drug validation. Drug validation is a harder problem given the diversity of chemical space---a model that is largely successful on the drug validation problem can be used as a virtual screening tool for compounds. While it may appear the CCLE samples performed better with drug validation, this is an anomaly given at most two CCLE drugs appeared in that validation set.

    Considering the case of a classification problem is often more actionable than the regression case for drug discovery and virtual screening tasks. The specific task of the model would be to filter models molecules based on their expected performance on CCLs which can be done at massive scale and quickly on a GPU, where inference times for these models well outperform standard virtual docking libraries (approx. 5000 samples per second, less if featurizing molecule on the fly). While other techniques such as virtual docking and simulation experiments can perform similar tasks by rank-ordering large virtual libraries, we propose the screening results from these models' predictions can be another measure designers of CCL panels and initial drug datasets can employ. The use of these models in the creation of new CCL panels would hopefully be able to balance the distribution of responders versus non-responders further. Given the models are regressors, a cutoff can be selected both at training time for using the balanced data sampling strategy and used afterwards for a sensitivity detection cutoff (we used 0.5, for example). Type I error (false positives) is a more costly error in the precision medicine task where the assignment of an incorrect drug is a wasted treatment opportunity, while type II error (false negatives) is costly in the virtual library screening where missing a potential lead at an early stage is a costly error. Of course, both errors should be considered for either task, but we aim to highlight the diversity of models possible just by training a large collection of DNN models. By altering the regression cutoff to 0.3, one can achieve $>90$\% balanced accuracy on the validation screening problem. We believe the cost analysis of virtual screening models is beneficial to researchers, where the choice of model used in screening should come from the task at hand, not necessarily the best state of the art performance. Further work can be done to introduce uncertainty quantification into these models using techniques in \cite{lakshminarayanan2017simple}. 
  
    While the prior classical assessment viewed the input data as the most significant point of variance, our results indicate the parameterization of the deep learning problem itself to be the largest explanation of variance. Based on the results, we see the transition from classical machine learning to deep learning will require a shift in focus from pure-feature engineering to deep learning hyperparameter optimization. Although we did not test discordant feature sets (all genetic features were super sets of each other, and the drug features techniques were both descriptor and fingerprint methods), one would expect to see a large variance across them,  though the predictive signal seems stable across all input features. This implies better representations exist for drugs and cells, or perhaps the features we have selected are optimal and a smaller set can be created. 
    
    While the hyperparameter decision tree ranked non-feature related aspects higher than the cell and drug featurizations, the result from the percentile and ANOVA analysis should not be understated, attributing significant to use of SNP data. For instance, the quantile analysis shows a greater difference in MCC on cell validation for the inclusion of SNPs than the differences between MCC for the various model architectures; yet, the decision tree feature importance for the cell validation ranks model architecture as a more useful predictor than the inclusion of SNP frames. We believe this should be read a probability difference, rather than a discrepancy. The decision tree is attempting to model the entire distribution of models we trained, so the fact that the optimizer has a big impact is only due to the fact that SGD is more likely to not converge or converge to the optimum point without hyperparameter optimization while Adam is likely to converge to the same point. In a sense, for researchers getting into deep learning, time spent understanding hyperparameters of deep neural networks has a higher probability of paying off than feature engineering initially. The quantile analysis, however, does indicate there are pay-offs to exploring feature sets, only if one is willing to pay the computational cost of hyperparameter sweeps or optimization.
    
    The results from this study indicate the choice of handling batch effect, general architecture, drug descriptors and modalities, and even the genetic features provided do not seem to impact the performance of the best models. The case of model architecture, dropout, and optimizers this result is not unexpected, as in general optimizers and dropout strategies should have minimal effect on the overall model capacity. Given the various model architectures were designed to have similar number of parameters, it seemed they were not differentiable in the case of general well performance. Batch effect handling, for example, employs a standard neural network to reduce the batch effect in the case of the combat scores; though this would be possible to occur within the first few layers of the network. To check, we present the by study scores of the models in the validation set split by the batch effect standardization procedure (Table \ref{tab:cross_scale}) which again do not seem to indicate a noticeable effect. Other aspects such as dropout and optimizer do not have a global effect on models as we trained many models, however there are smaller effects such as the Adam optimizer generally performing better than SGD but this may be due to the lack of global parameter optimization, adjusting factors such as batch size and learning rate as it is well studied that SGD requires careful selection of learning schedule \cite{zeiler2012adadelta}. 
    
    As a minor point, the choice of RNA-seq genes to use in the model was not an important consideration for model variance. There is that no superset of LINCS1000 gene set clearly outperformed any other set. This is, however, by design of the LINCS1000 gene set, as the selection of genetic features was created to recover most information contained in the transcriptome \cite{keenan2018library}. Given the ability of the network to learn at such a high drop out rate, the learned weights should have optimal sparse networks. We have shown that an optimal random subset of RNA-seq sizes is desirable. 
    
\begin{figure}[p]
            \centering
            \includegraphics[scale=0.7]{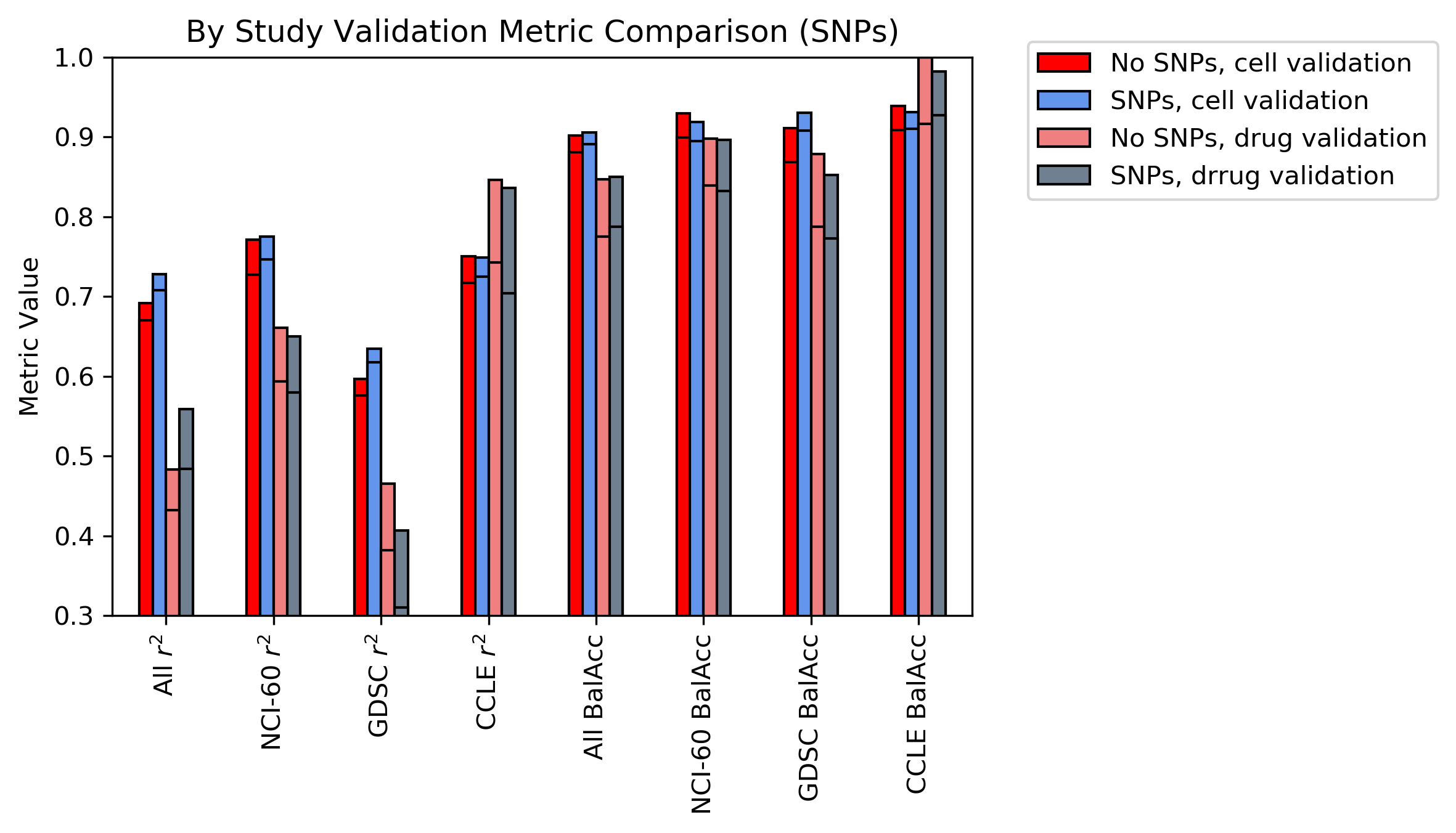}
            \caption{Cross study analysis of SNPs, 98th percentile and max for each score.}
            \label{fig:snpscross}
        \end{figure}

    The SNPs improved performance all around, when looking at aggregated test statistics. The results suggest that SNPs improved regression metric performance as well as classification metric performance, regardless of the validation strategy. It is observed that the NCI-60 cell line show over 100x mutation rates over other sources used. In order to examine if this artifact affected the results, we again break down the performance by cross study. We see that breakdown by study in the analysis further illustrates complexities around this type of study. The results in figure \ref{fig:snpscross} indicate however the inclusions of SNPs do not increase the performance across every study when aggregated from validation data. The cross-study results indicate that the larger mutation profile of the NCI-60 is not skewing the result indicating SNPs boost model performance on the cell validation split. 
        
    Both MOrderd and Dragon7 are fingerprint and descriptor based methods, with a great overlap between the two in terms of chemiformatics information found. An interesting component is the discrepancy between the cell and drug validation measures. When broken down by cross study, GDSC predictive performance is hurt severely when validating on drugs, and general predictions on unseen drugs is less powerful than unseen cells. We believe this is an indication that the models are not learning a useful representation of drugs, and further analysis of different drug modalities should be undertaken, such as images and fingerprints in \cite{cortsciriano2019}.

    \begin{table}[p]
\centering
\caption{Breakdown of data standardization techniques across the different studies in the validation data.}
\label{tab:cross_scale}
\begin{tabular}{@{}llllllllllll@{}}
                         &               &  & \multicolumn{4}{c}{Balanced Accuracy}            &  & \multicolumn{4}{l}{$r^2$ Score} \\ \midrule
validation               & scaling       &  & \multicolumn{1}{c}{All} & CCLE  & GDSC  & NCI-60 &  & All    & CCLE  & GDSC  & NCI-60 \\ \cmidrule(r){1-2} \cmidrule(lr){4-7} \cmidrule(l){9-12} 
\multirow{3}{*}{On Cell} & Combat        &  & 0.893                   & 0.914 & 0.912 & 0.902  &  & 0.708  & 0.729 & 0.619 & 0.749  \\
                         & None          &  & 0.893                   & 0.918 & 0.906 & 0.904  &  & 0.703  & 0.73  & 0.617 & 0.743  \\
                         & Source Scaled &  & 0.891                   & 0.91  & 0.904 & 0.899  &  & 0.71   & 0.727 & 0.617 & 0.747  \\
\multirow{3}{*}{On Drug} & Combat        &  & 0.797                   & 0.945 & 0.799 & 0.854  &  & 0.484  & 0.762 & 0.378 & 0.607  \\
                         & None          &  & 0.795                   & 0.944 & 0.794 & 0.854  &  & 0.47   & 0.761 & 0.388 & 0.598  \\
                         & Source Scaled &  & 0.796                   & 0.946 & 0.808 & 0.839  &  & 0.501  & 0.748 & 0.397 & 0.602  \\ \bottomrule
\end{tabular}
\end{table}

    \begin{table}[p]
\centering
\caption{Individual metrics for cell split validation for two of most useful models. The regression model is a differential dropout model with an initial dropout rate of 0.45, trained with the top 21 cancer type samples using loss weighting on the samples. Cells were featurized with SNPs and RNA-seq from the LINCS1000 subset, and chemicals were featurized by Dragon7 descriptors. The classification model is trained only on the top 6 cancer types and is the standard MLP model with a dropout rate of 0.2. Cells were featurized with SNPs and RNA-seq from the LINCS1000 subset, and chemicals were featurized by MOrdred descriptors. Both models used the Adam optimizer.}
\label{tab:cross_study}
\begin{tabular}{@{}llllllllll@{}}
      & \multicolumn{4}{c}{Best Regression Model}      &  & \multicolumn{4}{c}{Best Classification Model}  \\ \midrule
      & $r^2$ score & RMSE  & Balanced Accuracy & MCC  &  & $r^2$ score & RMSE  & Balanced Accuracy & MCC  \\ \cmidrule(r){2-5} \cmidrule(l){7-10} 
All   & 0.73        & 0.078 & 0.69              & 0.47 &  & 0.56        & 0.104 & 0.91              & 0.59 \\
CCLE  & 0.70        & 0.086 & 0.71              & 0.53 &  & 0.68        & 0.087 & 0.90              & 0.69 \\
GDSC  & 0.57        & 0.098 & 0.74              & 0.53 &  & 0.52        & 0.106 & 0.87              & 0.60 \\
NCI60 & 0.76        & 0.075 & 0.68              & 0.45 &  & 0.56        & 0.103 & 0.92              & 0.58 \\ \bottomrule
\end{tabular}
\end{table}   
 
 \section{Conclusion}
 We performed a large-scale sweep over feature types and training strategies for drug response prediction models. In doing so, we present a technique of training and testing across various cancer cell line screens. Despite the number of models trained and evaluated, we only comment on cancer drug response models using RNAseq, SNPs, and chemical descriptor methods. At this scale, we argue there is enough statistical (and observed) variance  for researchers creating models without the ability to perform large-scale hyperparameter optimization and architecture sweeps to take into consideration. Of course, these results may disagree with the finding of others as one-off model optimizations will create different results, as we base our results on $99^{\text{th}}$ percentile rather than one-off model scores. Further hyperparameter optimization should take place from this starting point. We found the inclusion of SNPs to be a significant factor for model performance. Contrary to previous studies, we do not find features to be the largest explanation for variance across models. We believe further work must done both in terms of architecture searches, data standardization, and evaluation across multiple data sources.
 
 \section*{Method} 
\subsection{Data}
    In this work, we curated a collection of cancer cell line screens from four different data sources. We use the Genomics of Drug Sensitivity (GDSC) \cite{Yang:2012d62}, NCI-60 Human Tumor Cell Lines Screen  (NCI-60) \cite{grever1992}, the Cancer Cell Line Encyclopedia (CCLE) \cite{Barretina:2012d62}, Cancer Therapeutics Response Portal (CTRP) \cite{klijn2014}, and the Genentech Cell Line Screening Initiative (gSCI) \cite{rees2015}. Each dataset consists of a panel of cancer cell lines tested of drugs. The datasets span size and specificity, as the NCI-60 contains the widest set of drugs, and the smallest set of cells. The datasets individually overlap both on some cells and drugs.
    
\begin{figure}[h]
    \centering
    \includegraphics[width=8cm]{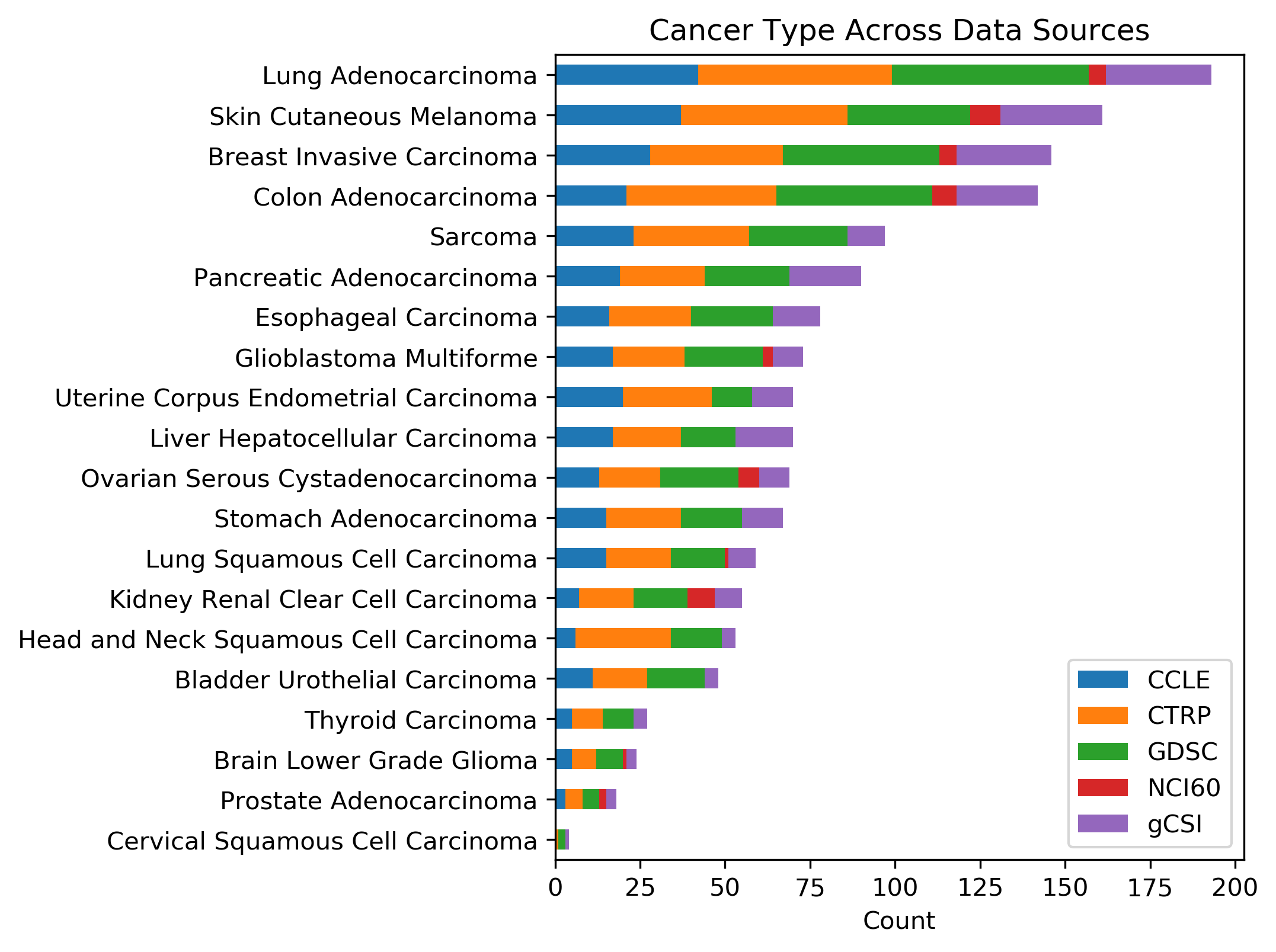}
    \label{fig:cancer_type_counts}
    \caption{Counts of cells based on type from each dataset used in the training data. Each data source contains other types not included, but we limit ourselves to the top 21 and top six cancer types. The types were determined based on clustering of RNA-seq using the autoencoder clustering method. While the method is not the same as a somatic tissue type or diagnosis, the type provides an indication of the diversity between data sources when attempting to balance data across classes.}
    \centering
    \end{figure}

    Cancer types across the datasets were not reported uniformly and contained some missing data. Using an autoencoder on RNA-seq data from the Genomic Data Commons \cite{grossman2016toward}, we were able to generate type-like labels for all RNA-seq data in our uniform dataset. Due to the imbalanced representation of various cancer types, the data was limited to the 21 most prevalent cancer types represented in the combined data frames. We applied a standard cancer type clustering technique to determine this source similarity metric (figure  \ref{fig:cancer_type_counts}) \cite{Needed}.

\subsection{Data Preprocessing}
    RNA-seq profiles are not provided in standard format by CTRP, gSCI, and GDSC; however, there is a mapping from cells in CCLE and NCI-60 to obtain RNA-seq profiles. After mapping and aligning to the set of drugs from each data frame, the cell response metric was standardized. The NCI-60 reports a dose-response metric based on cell growth after drug application. 

    \subsubsection{Cell Profile Processing}    
    The gene expression datasets for cancer cell lines, generated using RNA-seq and mutation data, were collected from the following sources: NCI-60, CCLE, and GDSC. The CTRP and gCSI drug response datasets were generated using the cell lines from CCLE dataset. Hence, for those, we used the gene expression and mutation data from the matching cell lines in the CCLE dataset. We refer to two featurization for cells, the RNA-seq for the expression data and single nucleotide polymorphisms (SNPs) for the mutation data. The SNP data was prepared as counts rather than binary indictor of presence or absence.
    
    The gene expression values were represented as fragments per kilobase of transcript per million (FPKM) values. To create varying gene lists, three datasets were created. The original contains a standard set of genes by an inner join of all the data sets (19,000 features), onco-gene set, and a LINCS1000 set. The LINCS1000 set was derived from the gene in The Library of Integrated Network-Based Cellular Signatures (LINCS) 1000 gene set \cite{koleti2017data}. The onco-gene set was created from a list of 2054 genes derived from the following three sources: i) 976 “landmark” human genes from high-throughput gene expression assay used in The Library of Integrated Network-Based Cellular Signatures (LINCS) 1000 study, ii) 470 high-confidence cancer genes identified in GDSC1000 study \cite{iori2016}, and iii) 1020 genes considered to be cancer genes by OncoKB \cite{wong2014}.
    
    The genes were filtered based on the genes int he respective dataset list and the FPKM values were transformed into log TPM values by$$\log \Big(\text{FPKM} \cdot 106 / \text{Sum of all FPKM values}\Big),$$ and batch effect procresssed according to a few different methods. 
    
\begin{figure}[h]
        \centering
        \includegraphics[width=\textwidth]{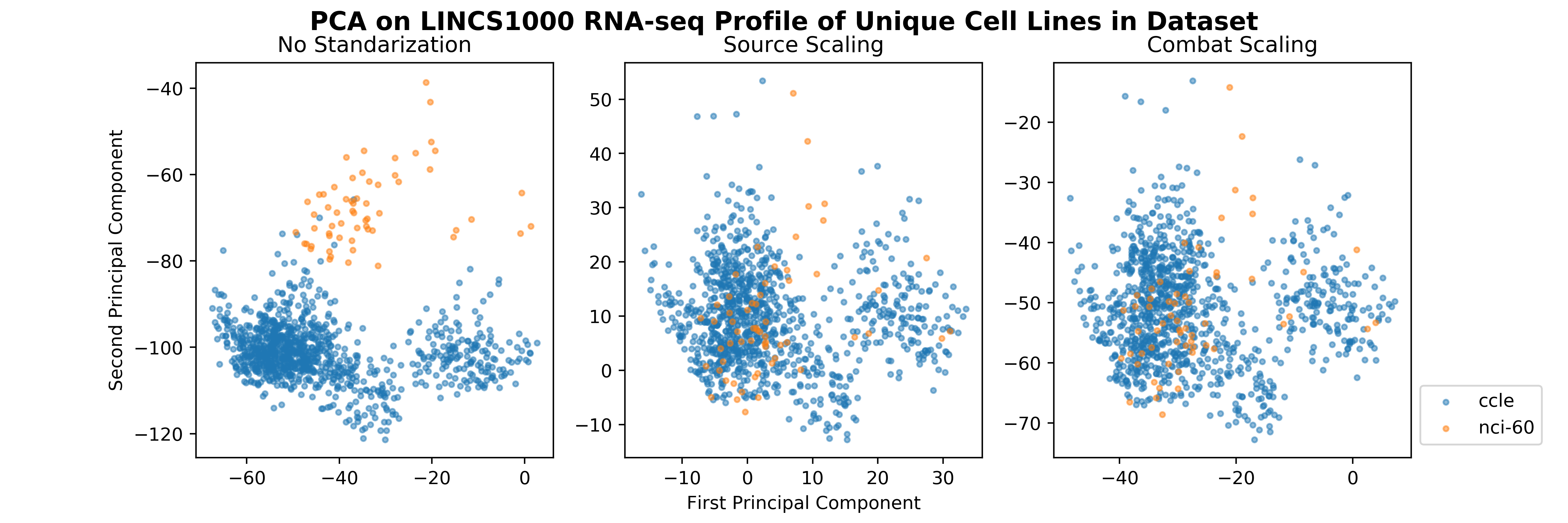}
        \caption{First two components from PCA on CCLE, GDC, and NCI-60 cell lines from our combined data frame. Without standardization, some batch effects are clear between NCI-60 and CCLE cell lines. The two methods tested seem to eliminate an obvious skew towards various batches. }
        \label{fig:pca_scaling}
    \end{figure}

     When batch effects are in the data frame, the biological signal is not as strong as the effect coming from the various batches (figure \ref{fig:pca_scaling}. This effects the downstream analysis such as differential expression and predictive modeling resulting in bias and unpredictable behavior. In order to manage batch effects between the different RNA-seq profiling, we tested three approaches: whole frame scaling, source scaling, and combat scores. \textit{Whole frame scaling}, or applying no batch effect handling method, merges the combined RNA-seq expression values and scaling each feature to unit norm. \textit{Source scaling} scales each feature to unit norm by the source rather than the combined data frame. \textit{Combat scaling} come from combat algorithm from Johnson \cite{johnson2007}.

    \subsection{Target Processing}
     Ideally, the target metric chosen correlates to growth inhibition used in the NCI-60 dataset for drug response measure given the NCI-60 is the largest data source used. The growth response data for NCI-60 cell lines is prepared by applying the compound at five different dosages to the tumor \cite{boyd_practical_1995}. By staining the cancer cells and measuring the absorbency with automated plate reader, they obtained the absorbency at time-zero, $T_z$, the control absorbency, $C$, and the absorbency after the application of the drug, $T_i$. Growth response used in the datasets is computed as Percentage Growth Inhibition (PGI) by
    \begin{equation} \label{eq:pgi}
    \text{PGI} = 
        \begin{cases}
            100\frac{T_i-T_z}{C-T_z} & T_i \geq T_z \\
            100\frac{T_i-T_z}{T_z}   & T_i < T_z
        \end{cases}.
    \end{equation}
    The PGI value is the target prediction value of most learning problems related to the NCI-60 dataset.
    
    From the PGI values across the different dosages, three values are computed related to the compound and cell. First, {Growth inhibition of 50\%} (GI50) is calculated from setting the PGI to 50\% and solving for $T_i$ in Equation \ref{eq:pgi}. The {Total Growth Inhibition} (TGI) is calculated by finding the concentration of the drug where $T_i=T_z$. Lastly the {50\% lethal concentration} (LC50) is the concentration of drug resulting in a 50\% reduction in the measured protein at the end of the drug treatment as compared to that at the beginning, which is the $T_i$ dosage so $(Ti-Tz)/Tz=-0.5 $. 
    
    However, this metric requires each sample is associated with a cell, drug, and a dosage. In order to remove the dependency on dosage as the interest in cell lines is skewed toward drug screening and further downstream precision medicine where dosage is not a preliminary question, we formulate a dose-independent drug response metric for the datasets provided by some data sources such as CCLE.  
    
    \begin{table}[p]
    \centering
  \begin{tabular}{@{}llll@{}}
\toprule
Source & $r^2$ fit       & $E_\infty$     & HS             \\ \midrule
CCLE   & $0.71\pm 1.35$  & $0.31\pm 0.35$ & $1.93\pm 1.50$ \\
CTRP   & $0.70\pm 0.36$  & $0.33\pm 0.38$ & $2.01\pm 1.53$ \\
GDSC   & $0.63\pm 0.40$  & $0.40\pm 0.39$ & $1.88\pm 1.53$ \\
NCI-60  & $0.80\pm 43.16$ & $0.36\pm 0.36$ & $1.93\pm 1.47$ \\
gCSI   & $0.82\pm 0.31$  & $0.34\pm 0.33$ & $1.90\pm 1.28$ \\ \bottomrule
\end{tabular}
    \label{table:dose_fit}
    \caption{Dose independent fitting results. $E_\infty$ and HS are parameters from the fit, and $r^2$ is from the result of the hill curve fit on a per-drug , per-cell basis. The NCI-60's large standard deviation comes from a few extreme outliers that were removed.}
    \end{table}
    
    For each dataset, we fit the dose dependent response to a hill curve with $E_\infty$, EC50, and hill sloop binding cooperatively (HS) (table \ref{table:dose_fit}). From the fit curve, we computed AUC - area under growth curve for a fixed dose range between 4 and 10 $-log_{10}(M)$, IC50 - drug dose to have 50\% growth, EC50se - standard error of the estimated EC50, R2fit - R2 score between the unclipped, real growths and fitted growth values, AUC1 - area under growth curve for the measured dose range in a study, AAC1 - area above growth curve for the measured dose range in a study, and DDS1 - drug sensitivity score \cite{yadav2015}.

    While all the metrics are highly correlated, in this paper, we study AUC for a fixed dose range across all studies from 4 to 10 $-\log_{10}(M)$. A value of 1 for AUC indicates a cell does not respond at dosages while a value of 0 indicates a cell responds completely at all dosages. Given most cell lines do not respond to a particular drug, the predictive target AUC is highly skewed (Figure \ref{fig:feat_Important}). 
        
\subsection{Molecular Representations}

   The representation of molecules plays a crucial role in response prediction for it directly indicates what kind and how much information will be given into the deep learning models during training and inference.
    
    We can divide the representations into engineered (knowledge-based) or non-engineered ones based on whether the featurization process involves the domain knowledge on molecular chemistry; alternatively, from the learning and data presentation point of view, the molecular representations can be either euclidean (vectors, tensors, voxels, etc.)or geometric (meshes, graphs, point clouds, etc.) ones.
    
    The most basic but yet still effective representation of molecules is SMILES string. Introduced in \cite{anderson1987smiles} in the late 1980s, SMILES (simplified molecular-input line-entry system) converts molecules into unique and human-understandable ASCII strings based on a given set of rules, which can be then processed with the methods and models in natural language processing. In our experiment, we used the canonical SMILES string, encoded them based on characters, then fed the vectors into deep learning models. Successful results has been demonstrated \cite{goh2017, kwon2017, hirohara2018} with this approach despite the fact that this vector representation requires minimal domain knowledge.
    
    Molecular fingerprint is a widely used method of encoding molecules based on the presence or absence of particular substructures. There are different ways to implement such substructure encoding, some of them are non-exhaustive such as MACCS, which only has 166 structural key, while the others are exhaustive but differs in searching patterns (circular-, path-, or tree-based searching) and substructure specifications (such as the number of atoms in substructure). In our experiment, we used ECFP\cite{rogers2010}, one of the most commonly used fingerprint, with different substructure size and vector dimensions to search for the most effective ECFP features for drug response.
    
    Molecular descriptor is a broad definition, which usually requires more domain knowledge and feature-engineering to generate. Generally speaking, any numeric representation that can "describe" the molecules in a certain way, is a descriptor. For the sake of better performance, a good set of descriptors often covers a wide range of molecular properties, some of which are rather complex and demands understanding of chemistry at a very deep level to design. In our experiment, we have tried Mordred \cite{moriwaki2018} and Dragon descriptors, both of which are highly popular and widely used in other works. In this work, we limit the exploration to molecular descriptors from MOrdered and Dragon7 code bases.

\subsection{DNN Models}
    Unlike classical algorithms studied in previous sweeps over this problem, deep neural networks are highly parameterized over architecture and training strategy \cite{zou2018primer}. A study of greater scale could be done just exploring the question of architecture and training strategy for a single set of features and data. Arbitrarily, we selected three model architectures based on experience from training models on the data, with the hope the three models are different enough to capture any interesting variance between them.
    
    We present three model architectures, a deep model, a model with a differential dropout scheme (inspired by the observations from \cite{zheng2014improving}), and a model with a sigmoid channel gated attention mechanism. The standard deep model is a simple ReLU based multi-layer perceptron (MLP). The differential dropout model is the same model except the dropout rates decrease to zero toward the output layer. The multiplicative channel gating is a two tower MLP, one for the genetic features of the cell and one for the drug features. The towers are combined using a multiplicative channel gate. 
    
    \subsubsection{Multiplicative Channel Gating}
    Attention mechanisms are used successfully in natural language processing (NLP) applications. Attention layers in NLP take as input a set of keys, values, and queries and attend to certain parts of the sequence to highlight important parts of the prediction \cite{vaswani2017}. Image attention expands the use case to images. Image attention has been used for image captioning where the model “attends” to a region in the image to predict word used to caption that part of the image \cite{xu2015}. We consider a self-attention mechanism between linear layers of a network, where an activation function (sigmoid, tanh, softmax, etc.) is applied to one channel and multiplied onto another, effectively attending to certain values of the activation. We employ a variety of generic activation functions such as sigmoid, tanh, or softmax. Unlike a standard self-attention implementation such as the one used in Monica et. al \cite{manica2019towards}, we do not enforce the use of softmax which can dimension the value of the gradients for non-sequence data. By using a sigmoid or tanh function, each channel is gated with the possibility of decreasing the value, though there is no restriction on how much information passes through the layer, unlike a softmax function.

    \subsubsection{Model Training}
    Three training methods were used across all models. Training was performed using Pytorch \cite{paszke2017automatic}. All models were trained on Summit, where each model used a single NVIDIA Volta V100. The models were dispatched for training and scoring using the Cancer Distributed Learning Environment (CANDLE) Supervisor \cite{wozniak2018}. Two optimizers, stochastic gradient descent (SGD) and Adam optimizer were tested with the learning rate initially set to 8e-4 for both\cite{kingma2014}. The learning rate was reduced by a quarter when the validation loss did not decrease for over 20 epochs. The batch size was not varied and set to 512. Each model was set to train for 400 epochs, and stop early if validation loss stopped decreasing after 10 epochs. We trained using Huber loss (Smooth L1 loss) with $\delta=1$ \cite{huber1964}.
    
    Besides the standard vanilla training procedure, two other training variations were used. Due to the high imbalance of positive leads (AUC $\leq 0.5$), an imbalanced data sampling strategy was used for each batch or a weighted loss function \cite{byrd2018, lauron2016}.  Imbalanced data sampling fills each batch with a balanced number of responders and non-responders. The drawback of this method is the model sees some data more than others which may lead to over-fitting on the smaller class. Loss weighting is generally used for classification where different weights are applied to different target predictions. As the models in this study are trained for a regression task, loss weighting for this context involves multiplying the loss of each batch by $1-y$, the target, in order to produce stronger gradients for non-responders and weaker gradients for responders---the idea coming from the the responders, having few examples, not being able to influence the gradient enough to push the model towards learning the smaller class. Neither strategy was tested in conjunction with the other.
    \subsubsection{Model Training}
    Three training methods were used across all models. Training was performed using Pytorch \cite{paszke2017automatic}. All models were trained on Summit, where each model used a single NVIDIA Volta V100. The models were dispatched for training and scoring using the Cancer Distributed Learning Environment (CANDLE) Supervisor \cite{wozniak2018}. Two optimizers, stochastic gradient descent (SGD) and Adam optimizer were tested with the learning rate initially set to 8e-4 for both\cite{kingma2014}. The learning rate was reduced by a quarter when the validation loss did not decrease for over 20 epochs. The batch size was not varied and set to 512. Each model was set to train for 400 epochs, and stop early if validation loss stopped decreasing after 10 epochs. We trained using Huber loss (Smooth L1 loss) with $\delta=1$ \cite{huber1964}.
    
    Besides the standard vanilla training procedure, two other training variations were used. Due to the high imbalance of positive leads (AUC $\leq 0.5$), an imbalanced data sampling strategy was used for each batch or a weighted loss function \cite{byrd2018, lauron2016}. Imbalanced data sampling fills each batch with a balanced number of responders and non-responders. The drawback of this method is the model sees some data more than others which may lead to over-fitting on the smaller class. Loss weighting is generally used for classification where different weights are applied to different target predictions. As the models in this study are trained for a regression task, loss weighting for this context involves multiplying the loss of each batch by $1-y$, the target, in order to produce stronger gradients for non-responders and weaker gradients for responders---the idea coming from the responders, having few examples, not being able to influence the gradient enough to push the model towards learning the smaller class. Neither strategy was tested in conjunction with the other.
    
\subsection{Model Evaluation}
    Prior works have utilized strict and non-strict partitioning for both cell lines and drugs in the training and testing set. Manica et al. marks the distinction between a strict and lenient split by cell and drug identity in the training set \cite{manica2019towards}. Chang et al. used a leniant (random on both cells and drugs) split for the validation data \cite{chang2018cancer}. Given our dose-independent prediction target, a specific cell and specific drug constitutes a single training example only in terms of model training and testing.
    
    Outside of pure model performance testing, there are two use cases we targeted: drug screening and precision medicine screening. For drug screening, panels are often performed against the set collection of cell lines to determine if a drug should move on to PDX models for further testing. In this case, we partition the training and test set by unique molecules, and use a 3-fold cross validation where each fold consist of entirely different molecules. For precision medicine, there is a list of approved drugs or known agents and the medical question is regarding which treatment to offer to a cell. In this case, the 3-fold cross validation is done over the unique cell lines, where each fold consists of unique cells not in any other fold. For the sake of testing more models and feature combinations, we did not perform a completely strict model evaluation of unseen drugs and cells as the use case is outside the scope of this research.  
    
    While this problem is posed as a regression problem as the AUC values are continuous on $0<AUC\leq 1$, viewing the problem as classification links it directly to application of the model. Due to the extreme skew of the training/testing distributions, $r^2$ and root mean squared error (RMSE) may not represent whether or not the model performs well in the region of interest (the model may minimize residuals over a large mass of the data, ignoring the residuals separating interesting response from no response). By artificially selecting a cutoff, 0.5, we can determine for the case of drug screening or precision medicine whether or not the model is learning to distinguish a responder cell/drug or a non-responder. The balance between type I and type II error is a calculation that is use-case specific and the selection of the feature set and model from our sweep will require consideration of this trade-off. We further justify this unusual approach to classification by the result from Jang \textit{et. al} indicating discretized target variables created less performant models from \cite{jang2014systematic}

\subsection{Analysis of Results}
    In order to test thoroughly the various proposed feature and training combinations, we tested the full combination space. Given 198 feature sets entailing various scaling, cell profile types, and drug representations, we ran each feature set on each model, with three different training strategies, top six and top 21 datasets, and three cross validation folds. In total, we ran 35,320 models on Summit. We gathered the results, and concatenates the predictions on each training fold from the model to estimate average statistics for the model.
    
    At this scale of models, some models (15\%) did not converge or provide reasonable results. Incorporating those failures into the overall performance of a model would not encapsulate the possible performance given hyper-parameter tuning and detailed work with a given configuration. We present models ranked by balanced accuracy and MCC to illustrate classification performance with post-processing binning and $r^2$ scores to illustrate the predictive power of the regression view of the problem.
    
    After gathering data, we removed models we believed to have not converged or over fit based on having one of the following criteria: TPR or FPR of 1.0, negative or zero $r^2$ score, or a loss outside the 85th quantile. This removed 2045 models from the data for analysis.
    
    In the results section, we highlight comparisons between the drug validation models and the cell validation models. The drug validations did not perform nearly as well as the cell validation models. We believe drug validation is a harder test on the model as the test is based on drug discovery rather than preclinical screening for a fixed set of drugs. For metrics such as Mathew correlation coefficient (MCC) and $r^2$ score we report 99th percentile, and for root mean squared error we report (0.01) percentile. We report multiple metrics as different purposes may arise for these models. We chose to report at 99th percentile versus best models found, as it is unclear with limited resources a researcher could reproduce the best model, and we aim to discuss approaches to the problem that can be taken out of the box and be performant.

 \section*{Acknowledgement}
This work has been supported in part by the Joint Design of Advanced Computing Solutions for Cancer (JDACS4C) program established by the U.S. Department of Energy (DOE) and the National Cancer Institute (NCI) of the National Institutes of Health. This work was performed under the auspices of the U.S. Department of Energy by Argonne National Laboratory under Contract DE-AC02-06-CH11357, Lawrence Livermore National Laboratory under Contract DE-AC52-07NA27344, Los Alamos National Laboratory under Contract DE-AC5206NA25396, and Oak Ridge National Laboratory under Contract DE-AC05-00OR22725.

This project has been funded in whole or in part with federal funds from the National Cancer Institute, National Institutes of Health, under Contract No. HHSN261200800001E.  The content of this publication does not necessarily reflect the views or policies of the Department of Health and Human Services, nor does mention of trade names, commercial products, or organizations imply endorsement by the U.S. Government.
    
\bibliographystyle{unsrt}  
\bibliography{references}

\end{document}